# Express Wavenet - a low parameter optical neural network with random shift wavelet pattern

Yingshi Chen[1]

[1] Institute of Electromagnetics and Acoustics, and Department of Electronic Science, Xiamen University, Xiamen 361005, China
E-mail: gsp@grusoft.com

**Abstract**

Express Wavenet is an improved optical diffractive neural network. At each layer, it uses wavelet-like pattern to modulate the phase of optical waves. For input image with $n^2$ pixels, express wavenet reduce parameter number from $O(n^2)$ to $O(n)$. Only need one percent of the parameters, and the accuracy is still very high. In the MNIST dataset, it only needs 1229 parameters to get accuracy of 92%, while the standard optical network needs 125440 parameters. The random shift wavelets show the characteristics of optical network more vividly. Especially the vanishing gradient phenomenon in the training process. We present a modified expressway structure for this problem. Experiments verified the effect of random shift wavelet and expressway structure. Our work shows optical diffractive network would use much fewer parameters than other neural networks. The source codes are available at https://github.com/closest-git/ONNet.

1. **Introduction**

Diffractive deep neural network (DNet)[1,2,3,4] is a novel machine learning framework, which could learn from the modulation of optical wave propagation. DNet is composed of stacked layers. Each layer has many pixels, which is an independent modulation element. Formula (1) gives the modulation of each pixel $p$:

$$modulation\ _p = a_p exp(j\emptyset_p) \qquad (1)$$

where $a_p$ would change the amplitude of optical wave and $\emptyset_p$ would change the phase. As indicated in [4], the modulation is actually an activation function on optical transformation. In this paper, we only discuss the modulation of phase. Only phase modulator $\emptyset_p$ would be learned by the network and $a_p$ would always be 1. [1,2,3] treats each pixel as an independent parameter. We find these pixels could be arranged in a wavelet pattern. As shown in figure 1, pixels at the same wave circle have same $\emptyset_p$. So for a layer with $n^2$ pixels, wavelet pattern would reduce the parameter number from $O(n^2)$ to $O(n)$. For the case of a 10 layer network and each layer's side length is 112, the parameter number would be reduced from 125440 to 1229. Only 1% compared to the random pattern in [1,2,3].

The wavelet pattern not only reduces the number of parameters greatly, but also reveals more intrinsic characteristics of optical network. Observe the changing process of wavelet in training, we can see the obvious "vanishing gradient" phenomenon [12]. Only the wavelet in the last layer has significant changes. The wavelets in previous layers are almost the same as the initial pattern. In the case of deep convolutional neural networks(CNN)[13,14], the vanishing gradient problem is a long-standing problem and has been studied deeply. But in the case of optical neural networks, it has not been reported and analyzed. Out wavelet pattern would help optical network researcher to deal with this problem.

It's not easy to train a DNet with so few parameters, we present two novel techniques. As shown in figure 1, the first is shift wavelet technique, which translates the wavelet randomly. The second is expressway structure. The output of each layer is not only the input of the next layer, but also directly accumulated to the last layer with weighting coefficient. In the remainder of this article, we call this improved network with these two features as express wavenet(ExWaven). And experiments verified its effectiveness. In the MNIST dataset, it only need one thousand parameters to get accuracy of 92%, which is much less than the classical deep CNN. To our knowledge, it's maybe the minimal network for the MNIST classification problem.

For the simulation of express wavenet, we developed ONNet. It's an open-source Python/C++ package and the codes are available at https://github.com/closest-git/ONNet. ONNet would provide a lot of tools for researchers studying optical neural networks.

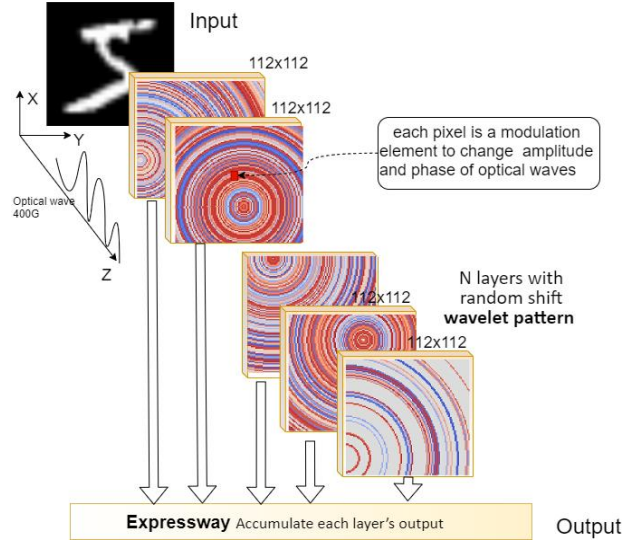
Figure 1 the structure of Express Wavenet.

## 2. Express Wavenet with random shift wavelet

Figure 1 showes the detailed structure of Express Wavenet. At each layer, the propagation of optical waves could be described by formula (2). For the detail information, please see [1,2,3,4].

$$\begin{cases} h_i^l = t_i^l z_i^l \\ z_i^l = \sum_k \omega_{k,i}^{l-1} h_k^{l-1} \\ t_i^l = a_i^l \exp(j\phi_i^l) \end{cases} \quad (2)$$

where

$\omega_{i,p}^l = \frac{p-p_i}{r^2}\left(\frac{1}{2\pi r} + \frac{1}{j\lambda}\right) \exp\left(\frac{j2\pi r}{\lambda}\right)$ are constants for the wave propagation

$a_i^l$ (amplitude), $\phi_i^l$ (the pahse value ) are learnable parameters

The key component of express wavenet is random shift wavelet and expressway network structure. The details are as follows.

### Random shift wavelet

Let $n$ is the side length of each layer, each pixel $p$ is represented by its coordinate as $(x,y)$, $x,y \in \{1,2,\cdots,n\}$. In the previous work[1,2,3], all $n^2$ pixles are independent variables as the formula (3).They take random initial value between $(0,2\pi)$. Then network would adjust the $\phi_p$ to learn a better model. So the total number of parameters is $n^2$.

$$\phi_p = \text{random}(0,2\pi) \text{ for each } p(x,y) \ x,y \in \{1,2,\cdots,n\} \quad (3)$$

We present a wavelet-like pattern to reduce the number of parameters. We first select a fixed point $q$ randomly. Then set $\phi_p$ on the distance between p and $q$. The detailed algorithm is as follows:

Algorithm 1 Set $\phi_p$ by the random shift wavelet

> 1 Randomly select a fixed point $q = (x_q, y_q)$
> 2 for each pixel $p(x_p, y_p)$ in the layer, get its $L^1$ distance to $q$
>   $L^1$ distance between p and q: $L(p,q) = |x_p - x_q| + |y_p - y_q|$
> 3 Find all concentric circles
> 4 For each circle $C$, set a random value $\phi_C \in (0,2\pi)$
> 5 For each pixel $p$ in circle $C$, $\phi_p = \phi_C \ \forall p \in C$

Figure 1 showes some typical pattern of each layer. It's clear that many concentric circles pixels just like wavelet on water surface. So we call these patten as wavelet. There are at most $\sqrt{2}n$ concentric circles at each layer, so the total number of paramter is reduced to O(n).

We could use wavelet pattern without random translation. That is, all fix points $q$ are the center of layer($q = (n/2, n/2)$). This will reduce accuracy. The experiment on different datasets showed this point clearly. Just like the training of depp CNN, randomness always improve the accuracy and make the learned model has better genelarity.

### Expressway network structure

"Vanishing gradient" is a common problem in the training of neural networks, especially for deep networks with many layers. There is a similar phenomenon in the optical diffractive networks. In the case of wavelet network, there are intuitive and vivid displays. As figure 4 shows: only the wavelet in the last layer has significant changes. The changes in the previous layers are much smaller.

There are many techniques for this problem in classical deep learning[23,24]. Formula (4) defines the transform used in the highway networks[23].

$$z^l = f(z^{l-1}, W^{l-1})T^l + z^{l-1}(1 - T^l) \qquad (4)$$

where $z^l$ is the output at $l$ layer, $W^l$ is the weighting matrix. $T^l$ is the gating function

In this propagation process, current layer would directly use some output of previous layer. In some sense, this creates an information highway to pass gradient information. So highway networks would train very deep networks with hundreds of layers. But for the optical network, there's no easy way to implement the gating function. The output of each diffractive layer is actually optical waves at speed of light. For some diffractive layer in the middle, how accumulate previous layer's output? That means the stop the transmission of light at middle layer to merge the output of previous layer. We present a modified version of highway network. We just merge the output of middle layers at the final output layer. No merge operation in the middle layers. Only the final output is accumulated as the following formula.

$$z_{output} = \sum_l z_l w_l \qquad (5)$$

We observed similar phenomenon in the testing dataset. That is, these expressway structures would send gradient information to the previous layer and improve the accuracy.

## 3. Results and discussion

We tested express wavenet(ExWave) on two datasets MNIST and fashion-MNIST[21]. MNIST database is a commonly used database of handwritten digits, which contains 60,000 training images and 10,000 testing images. fashion-MNIST consists of a training set consisting of 60000 examples belonging to 10 different classes and a test set of 10000 examples. Figure 2 shows some pictures of MNIST and fashion-MNIST. As indicated in [4], optical diffractive network still in its infancy. So these two simple testing datasets can still be used for testing. We would test larger and harder datasets in the later works.

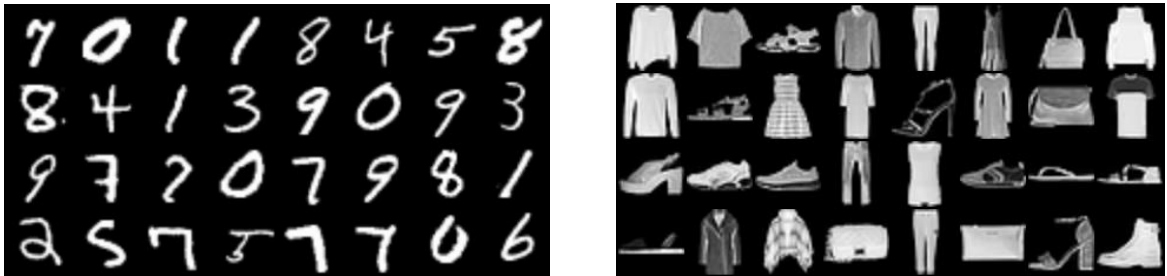

Figure 2 MNIST and fashion-MNIST datasets (10 classes)

We trained a 10-layer ExWave with the side length of each layer is 112. This network has only 1229 parameters. The parameter numbers in each layers are (130,118,112,143,126,104,141,94,108,143). Figure 3 plots the learning curves on two testing datasets. The classification accuracy is 92% on the MNIST and 81% on the fashion-MNIST. We also tested different algorithm combinations to compare performance. ExWave has two unique techniques. One is shift wavelet technique. The other is expressway structure. So there are four combinations, which listed in table 1. It's clear that both techniques can significantly improve accuracy. Compared to fixpoint wavelet, the random translation improve the accuracy from 80.3% to 90.8% in MNIST and from 71.7% to 78.7% in fashion-MNIST.

Table 1 Accuarcy on the different algorithm combinations

|  | MNIST | fashion-MNIST |
|---|---|---|
| Express Wavenet | 92.3% | 80.9% |
| Only shift wavelet | 90.8% | 78.7% |
| Only Expressway | 89.3% | 77.4% |
| No shift wavelet;No Expressway | 80.3% | 71.7% |

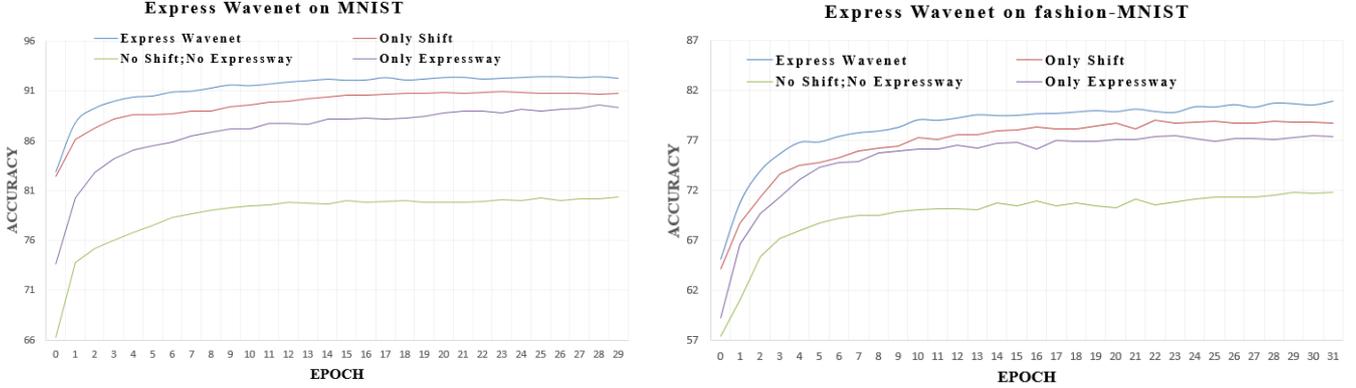

Figure 3 Comparison of Express Wavenet on MNIST and fashion-MNIST.
Four curves correspond to four algorithm combinations. The blue curve corresponds to "Express Wavene". The red curve corresponds to network with only shift wavelet. The violet curve corresponds to network with only expressway. The cyan curve corresponds to an trivial netwok.

The wavelet pattern of express wavenet provides a vivid way to reveal the detail of training process. Figure 4 shows the wavelet pattern at different epoch. These images actually show the sine transformed $\emptyset_p$. No matter the value of $\emptyset_p$, $a_p$ would always be in the range of $[-1,1]$.

$$a_p = \sin(\emptyset_p) \quad \emptyset_p \in R \text{ and } a_p \in [-1,1] \tag{6}$$

The first row is the initial random pattern. There are dense concentric circles with many mutations. As the training goes on, there are obvious changes at the last layer, at which the wavelet becomes sparse and more regular. The pattern at the previous layers is also changing, but it's much smaller. It's a clear sign of "vanishing gradient" phenomenon.

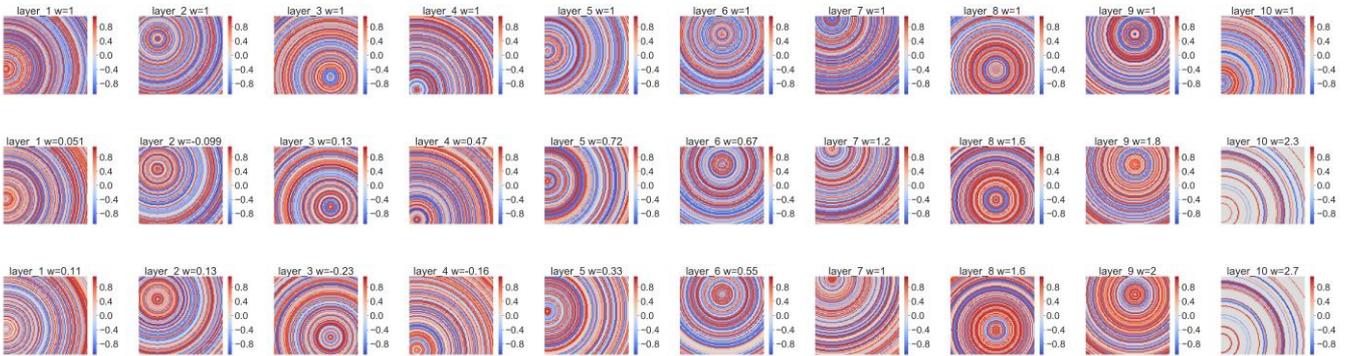

Figure 4 The phase maps of 5 layers at different epoch.
The first row shows wavelet pattern at epoch 0. The middle row shows wavelet pattern at epoch 10. And the last row shows wavelet pattern at epoch 20.

## 4. Conclusion and prospect

In this paper, we present express wavenet, which reveal more characteristics of optical diffractive networks. First, diffractive network has the potential to use much fewer parameters than other networks. By our random shifted wavelet, only one percent of the parameters are needed. Second, there are obvious vanishing gradient phenomenon in the training process. The classical method in the deep CNN for vanishing gradient problem is hard to realize with optical device. So we present expressway structure. But the changes in the previous layers are still small. The expressway is only the first step. We would study the problem further in later work.